\documentclass[11pt]{article}

\usepackage{acl}

\usepackage{times}
\usepackage{latexsym}
\usepackage[T1]{fontenc}
\usepackage[utf8]{inputenc}
\usepackage{microtype}
\usepackage{inconsolata}
\usepackage{graphicx}
\usepackage{url}
\usepackage{booktabs}
\usepackage{iftex}
\usepackage{tikz}
\usepackage{pgfplots}
\pgfplotsset{compat=1.18}

\newcommand{\rankmedalicon}[3]{  \tikz[baseline=-0.6ex, x=1ex, y=1ex]{    \fill[blue!70!black] (-0.60,1.00) -- (-0.18,0.30) -- (-0.42,0.30) -- (-0.82,1.00) -- cycle;
    \fill[blue!50!cyan] (-0.42,1.00) -- (-0.05,0.30) -- (0.08,0.30) -- (-0.02,1.00) -- cycle;
    \fill[blue!70!black] (0.60,1.00) -- (0.18,0.30) -- (0.42,0.30) -- (0.82,1.00) -- cycle;
    \fill[blue!50!cyan] (0.42,1.00) -- (0.05,0.30) -- (-0.08,0.30) -- (0.02,1.00) -- cycle;
    \fill[#1] (0,0) circle (0.75);
    \draw[black!55, line width=0.08ex] (0,0) circle (0.75);
    \node[font=\bfseries#3, text=white] at (0,0) {#2};
  }}
\newcommand{\rankmedal}[1]{  \makebox[2.4em][c]{    \ifnum#1=1
      \rankmedalicon{orange!80!yellow}{1}{\fontsize{5}{5}\selectfont}    \else\ifnum#1=2
      \rankmedalicon{black!30}{2}{\fontsize{5}{5}\selectfont}    \else\ifnum#1<6
      \rankmedalicon{brown!70!orange}{#1}{\fontsize{5}{5}\selectfont}    \else\ifnum#1<10
      \rankmedalicon{black}{#1}{\fontsize{5}{5}\selectfont}    \else
      \rankmedalicon{black}{#1}{\fontsize{5}{5}\selectfont}    \fi\fi\fi\fi
  }}
\newcommand{\rank}[1][]{  \rankmedal{#1}}

\title{ClinicalEncoder26AM: A Multlilingual Diagnosable ColBERT Model; \\ Evidences from the MultiClinNER Shared Task}

\author{François Remy \\
Parallia AI \\
\tt{francois.remy@parallia.eu}}

\begin{document}
\maketitle

\begin{abstract}
ClinicalEncoder26AM is a multilingual Diagnosable ColBERT \citep{diagnosable_colbert_2026} for clinical and biomedical texts, which aligns at multiple levels its token-level semantic with ClinicalMap25, a clinical latent space inspired by BioLORD-2023 and enriched with synthetic and annotated supervision. 
The post-training recipe builds upon BGE-M3 \citep{chen-etal-2024-m3}, and combines synthetic clinical notes, patient--doctor conversations, and annotated resources such as MedMentions \citep{mohan-li-2019-medmentions}, while considering both named-entity-level and sentence-level representations in a multi-adapter distillation, along with a ColBERT-style retrieval objective. 
In this system demonstration paper, we evaluate the model in the MultiClinNER shared task by finetuning it as a BIO tagger for patient symptoms, disorders, and procedure spans, using a lightweight two-layer CNN head to improve local boundary detection. The resulting system remains simple, processes most documents in a single 8192-token window, and achieves state-of-the-art multilingual entity recall, while achieving Top 5 overall across all entity types and languages in Character-weighted F1 scores. 
Training curves further show that ClinicalEncoder26AM is markedly more data-efficient than the base M3 model, supporting the usefulness of its clinical post-training for downstream information extraction.
\end{abstract}

\section{Introduction}

ClinicalEncoder26AM is the multilingual extension and improvement of the ClinicalEncoder series of interpretable clinical encoders designed to support both high-quality retrieval and token-level semantic inspection.
The model follows the Diagnosable ColBERT paradigm, in which each token is not only encoded for late interaction, but also aligned with a clinically grounded latent space that serves as a semantic reference during analysis and debugging \citep{diagnosable_colbert_2026}.
This design is motivated by the difficulty of evaluating the alignment between ColBERT retrieval models and expert judgment over concept similarities.
While ColBERT models are intrepretable at the document--query interaction level, their embeddings themselves remain opaque.
By contrast, a diagnosable model can expose whether a token or span is being mapped toward the appropriate region of a domain-specific semantic space, making failures easier to address.

ClinicalEncoder26AM builds on ColBERT-style late interaction, but adapts it to multilingual clinical and biomedical text through post-training on top of BGE-M3 \citep{chen-etal-2024-m3}.
Its semantic grounding relies on ClinicalMap25, a latent clinical space inspired by BioLORD-2023 \citep{remy2024biolord} and enriched with additional synthetic and annotated signals.
The supervision used for this post-training includes synthetic clinical notes, patient--doctor conversations, and annotated biomedical corpora such as MedMentions \citep{mohan-li-2019-medmentions}.
The overall goal is not only to produce strong multilingually-aligned representations, but also to retain a meaningful alignment between local token semantics, higher-level sentence information, and retrieval-oriented document representations.

In this paper, we focus on the MultiClinNER shared task \cite{multiclinai-overview-2026} of SMM4H-HeaRD \cite{smm4h-heard-overview-2026} rather than on the post-training details.
Our main objective is to describe how a clinically post-trained multilingual encoder can be adapted into a practical BIO tagging system for extracting symptoms, disorders, and procedures from clinical texts.
This downstream setting is particularly informative because it highlights an important distinction: even when token representations are rich enough to support ontology-grounded interpretation, they are not automatically optimized for precise span boundary detection.
In other words, knowing what a token means in context is not the same as deciding exactly where entities begin and end in running text.

For that reason, we finetune ClinicalEncoder26AM with a lightweight task-specific tagging head instead of attempting to recover spans directly from ontology mappings.
This choice keeps the system simple, benefits from the shared task annotations, and provides a clean way to evaluate whether the post-trained multilingual encoder transfers effectively to structured clinical information extraction.
As we show later, this simple approach yields a strong recall-oriented system across all languages, with competitive overall rankings and no major degradation in languages such as Czech that were not exercised in the original post-training.

\section{Methodology}

ClinicalEncoder26AM is built from the multilingual BGE-M3 encoder \citep{chen-etal-2024-m3}, whose backbone follows an XLM-RoBERTa-style architecture with 24 Transformer layers, a hidden size of 1024, 16 attention heads, and supports up to 8192 simultaneous tokens (about 32kB of raw English text).
In practice, this context size is sufficient to encode almost all documents in the MultiClinNER corpus a single forward pass.
Unlike generative language models, ClinicalEncoder26AM computes all document representations in one pass, making it suitable for both large-scale retrieval and clinical information extraction.

The central methodological idea is to couple late-interaction retrieval representations derived by down-projection from embeddings aligned with a clinically grounded reference space.
More precisely, each token is encouraged to align with ClinicalMap25, a latent semantic space designed to capture fine-grained clinical meaning at the token, mention, and local contextual levels.
ClinicalMap25 is inspired by BioLORD-2023 \citep{remy2024biolord}, but extends that line of work with additional supervision signals and more recent synthetic data sources in order to represent clinical concepts more precisely in actual clinical language.
This alignment is intended to make the model more diagnosable: instead of only inspecting token-to-token interaction scores after retrieval, one can also inspect whether the token embeddings themselves are organized in a clinically sensible way.

A mixture of synthetic clinical notes, patient--doctor conversations, and annotated biomedical documents such as MedMentions \citep{mohan-li-2019-medmentions} are used as part of this post-training, including curated translations.
These resources expose the encoder to both biomedical terminology and more natural clinical phrasing, including context-sensitive formulations that are common in real-world documentation.
During this stage, we consider supervision at multiple granularities, including named-entity-level representations and sentence-level representations, before applying a ColBERT-style late-interaction training objective.
This multi-granular training setup is meant to preserve local semantics while maintaining compatibility with document-level retrieval behavior.

A second important component is the use of multi-adapter distillation to combine complementary supervision signals into a single multilingual model.
Rather than relying on one homogeneous source of supervision, the post-training transfers knowledge from multiple task-specific or data-specific views into a unified encoder.
This strategy is particularly useful in the clinical domain, where no single dataset fully captures ontology grounding, contextual phrasing, retrieval behavior, and multilingual variation at once.
It also helps retain the multilingual alignment inherited from BGE-M3 while adapting the model for clinical semantics.

As a result, ClinicalEncoder26AM can be used in several closely related ways.
The same encoder can support multilingual retrieval, direct token-to-concept mapping against ontologies such as SNOMED CT \citep{donnelly2006snomedct} or UMLS \citep{bodenreider2004umls}, and downstream token classification after task-specific finetuning.
From a modeling perspective, this is possible because the post-training is not restricted to one output format: it jointly encourages clinically meaningful local embeddings, sentence-aware contextualization, and compatibility with ColBERT-style late interaction.
In the present paper, we focus on the final use case, namely transfer to BIO tagging for shared-task named entity recognition.

\section{Experimental Setup}

For the MultiClinNER shared task, we finetune ClinicalEncoder26AM as a BIO sequence tagger for symptom, disorder, and procedure extraction in seven European languages.
All target languages are supported by the multilingual base model, although only a subset was explicitly represented in the original post-training mixture.
Following the shared-task setup, we train one model per entity type rather than a single joint tagger, as result of the format of the annotation data.

The tagging architecture is intentionally lightweight.
On top of the encoder outputs, we add a two-layer CNN block with window size of 5 and hidden size of 1024 before the final token classification layer.
This design is motivated by the fact that span boundary detection depends strongly on local contextual cues, even when the underlying token representations are already semantically rich.
In practice, the convolutional block provides a simple way to aggregate short-range evidence around each token without departing from the base encoder architecture.

Training is performed on the MultiClinNER training split \citep{lima_lopez_2026_18508039} for 4 epochs using a standard token-level classification objective.
Inference then applies an argmax decision at each token and extracts valid spans directly from the BIO sequence by reading patterns of the form $B?I+$.
We also experimented with a CRF decoding layer on top of the token classifier, but it did not improve validation behavior and was therefore not retained in the final system.
This result is consistent with the overall design choice of keeping the downstream architecture simple and letting most of the semantic burden be handled by the post-trained encoder.

The long-context capacity of ClinicalEncoder26AM plays an important practical role in this setup.
Because the model can process up to 8192 tokens at once, nearly all documents in the MultiClinNER training data can be encoded in a single window.
This substantially simplifies multilingual preprocessing, avoids many sentence-splitting heuristics, and reduces the number of boundary errors introduced by document chunking.
Only a small number of pathological documents in the background portion of MultiClinCorpus required special handling (automatically translated documents containing degenerate loops that artificially inflated their length).

For those rare long documents, we apply a simple overlapping-window strategy.
Each document is split into partially overlapping chunks, token predictions are computed independently in each chunk, and the final label for a token is taken from the window that gives it the largest amount of surrounding context.
This heuristic is sufficient for the shared-task setting and avoids introducing more complex sequence-reconciliation machinery.
Overall, the experimental setup is deliberately conservative: rather than designing a sophisticated decoding pipeline, we evaluate how far a strong clinically post-trained multilingual encoder can go with a minimal and reproducible BIO-tagging head.

The shared task reports both strict and character-based evaluation metrics, each broken down into precision, recall, and F1. Strict metrics require an exact match between a predicted span and a gold span, and therefore strongly penalize even small boundary errors. Character-based metrics instead compare the overlap at the character level, which gives partial credit when a system identifies the right mention region but misses the exact start or end boundary. Because entity boundaries are always a bit arbitrary due to the inherent debatability of the inclusion or exclusion of concept modifiers, we focus on the character-level metrics.

\section{Results and Discussion}

\begin{table*}[t]
\centering
\small
\begin{tabular}{lcccccc}
\toprule
Entity Type & Char R & Char P & Char F1 & Strict R & Strict P & Strict F1 \\
\midrule
PROCEDURE & $0.85 \pm 0.02$ & $0.81 \pm 0.01$ & $0.83 \pm 0.01$ & $0.73 \pm 0.03$ & $0.69 \pm 0.02$ & $0.71 \pm 0.02$ \\
DISORDER & $0.85 \pm 0.02$ & $0.80 \pm 0.02$ & $0.82 \pm 0.02$ & $0.73 \pm 0.03$ & $0.69 \pm 0.03$ & $0.71 \pm 0.03$ \\
SYMPTOM & $0.80 \pm 0.03$ & $0.73 \pm 0.02$ & $0.76 \pm 0.02$ & $0.65 \pm 0.04$ & $0.59 \pm 0.03$ & $0.62 \pm 0.03$ \\
\bottomrule
\end{tabular}
\caption{Summary of the MultiClinNER results of our ClinicalEncoder26AM system (Character-weighted \& Strict Recall, Precsion, and F1; all reported as mean $\pm$ standard deviation across the 7 languages).}
\label{tab:main-results}
\end{table*}

The results on the private test set of the shared task demonstrated that our ClinicalEncoder26AM-based system performed exceptionally well, and proved competitive along the 6 ranking metrics across all entity types and languages, even outscoring larger systems and monolingually-finetuned models.
Across the shared-task submissions, it consistently achieves excellent recall, with best character-weighted recall performances in all languages except Spanish, and with top-3 positions in strict recall for a majority of language--entity combinations.
At the same time, its precision was systematically less highly-ranked than its recall, indicating that the system identifies a large fraction of relevant spans, but still leaves room for improvement in filtering spurious predictions and refining exact boundaries.
This pattern is visible in Table~\ref{tab:main-results}, where the results show substantially stronger character recall than precision for all three entity types.

In practice, this means that the model usually localizes clinically relevant evidence correctly, but sometimes breaks long mentions into multiple nearly-consecutive spans (hurting precision).
This is consistent with the model design: the post-trained encoder appears to provide broad semantic coverage, while the lightweight downstream head emphasizes transfer simplicity rather than aggressive decoding control.

A second important observation is that, in small-scale comparisons, multilingual finetuning appeared to work slightly better then language-specific training. We hypothesize that other languages function as a form of data augmentation.
Although this advantage is not large enough to justify a strong claim about universal superiority, it suggests that the multilingual alignment inherited from BGE-M3 remains useful after clinical post-training, as languages benefit from each-other training data.
This is particularly relevant for languages such as Czech, which were not part of the original post-training mixture but nevertheless do not exhibit a major degradation in downstream performance.
The absence of a clear low-resource collapse supports the idea that the shared multilingual semantic space of the base model remains coherent after adaptation to clinical domain.

The optimization behavior further supports the value of the post-training stage.
In training-loss comparisons against the base M3 model, ClinicalEncoder26AM converges faster and maintains a lower loss throughout most of the finetuning run, although the gap narrows toward the end (see Annex \ref{ann:training_loss}).
This suggests that the post-trained encoder starts from a representation space that is already better aligned with the shared-task objective, and therefore requires less supervision to reach strong token-classification performance.
From a practical perspective, this is an important result: it indicates that the additional clinical adaptation is not only semantically meaningful, but also directly beneficial for downstream data efficiency.

These results also clarify the relationship between ontology-aware semantic grounding and span extraction.
ClinicalEncoder26AM can directly map tokens toward ontology-linked concepts in resources such as SNOMED CT or UMLS, but this capability alone does not solve the problem of extracting precise mention boundaries in free text.

In principle, one could attempt a more direct ontology-driven extraction pipeline, but in our setting that approach proved more complex than a standard supervised BIO finetune and offered no significant advantage given the amount of labeled data available.
The present results therefore favor a pragmatic division of labor: use the post-trained encoder for strong multilingual semantic representations, then learn explicit span boundaries with a lightweight supervised head.

Finally, the error profile suggests an obvious direction for future improvement.
Because recall is already strong, the most promising gains are likely to come from better filtering, confidence calibration, or boundary-aware post-processing rather than from making the encoder broader or more sensitive.

\section{Conclusion}

In the MultiClinNER Shared Task, we demonstrated that a multilingual Diagnosable ColBERT grounded in ClinicalMap25 such as ClinicalEncoder26AM can transfer effectively to multilingual clinical named entity recognition with only a lightweight supervised head.
Despite the simplicity of the BIO-tagging setup, our system achieved the best recall across nearly all languages, and proved competitive on all ranking metrics.
Our clinical post-training delivered improvements in both training speed and downstream effectiveness.
These results suggest that clinically grounded, interpretable token representations are not only useful for retrieval analysis, but also provide a strong foundation for practical information extraction.
More broadly, the shared-task findings support a pragmatic perspective: ontology-aware semantic grounding and simple task-specific finetuning can complement each other well, and further gains are more likely to come from better filtering and boundary refinement of already-knowledgeable encoders rather than from the usage of substantially less effective large language models.

\newpage
\bibliography{custom}

\clearpage
\onecolumn
\appendix

\section{Appendix: Full Results Table}

\begin{table}[h!]
\centering
\scriptsize
\begin{tabular}{llcccccccc}
\toprule
Entity Type & Lang & Char R rk & Char P rk & Char F1 rk & Strict R rk & Strict P rk & Strict F1 rk & Char F1 & Strict F1 \\
\midrule
PROCEDURE & cz & \rank[1] & \rank[8] & \rank[3] & \rank[3] & \rank[7] & \rank[5] & 0.8276 & 0.6880 \\
PROCEDURE & en & \rank[1] & \rank[11] & \rank[5] & \rank[4] & \rank[11] & \rank[8] & 0.8348 & 0.7110 \\
PROCEDURE & es & \rank[2] & \rank[11] & \rank[8] & \rank[8] & \rank[11] & \rank[8] & 0.8476 & 0.7536 \\
PROCEDURE & it & \rank[1] & \rank[9] & \rank[3] & \rank[2] & \rank[8] & \rank[5] & 0.8275 & 0.6969 \\
PROCEDURE & nl & \rank[2] & \rank[10] & \rank[3] & \rank[3] & \rank[10] & \rank[5] & 0.8112 & 0.6922 \\
PROCEDURE & ro & \rank[1] & \rank[9] & \rank[3] & \rank[2] & \rank[8] & \rank[4] & 0.8396 & 0.7222 \\
PROCEDURE & sv & \rank[1] & \rank[8] & \rank[3] & \rank[3] & \rank[8] & \rank[5] & 0.8275 & 0.7015 \\
\midrule
DISEASE & cz & \rank[1] & \rank[10] & \rank[4] & \rank[3] & \rank[8] & \rank[5] & 0.7983 & 0.6737 \\
DISEASE & en & \rank[1] & \rank[13] & \rank[6] & \rank[7] & \rank[13] & \rank[9] & 0.8510 & 0.7520 \\
DISEASE & es & \rank[6] & \rank[10] & \rank[8] & \rank[8] & \rank[9] & \rank[8] & 0.8431 & 0.7512 \\
DISEASE & it & \rank[1] & \rank[11] & \rank[3] & \rank[3] & \rank[9] & \rank[8] & 0.8235 & 0.7027 \\
DISEASE & nl & \rank[3] & \rank[8] & \rank[3] & \rank[3] & \rank[9] & \rank[7] & 0.7926 & 0.6778 \\
DISEASE & ro & \rank[1] & \rank[10] & \rank[3] & \rank[3] & \rank[6] & \rank[3] & 0.8307 & 0.7189 \\
DISEASE & sv & \rank[1] & \rank[8] & \rank[4] & \rank[3] & \rank[6] & \rank[5] & 0.8058 & 0.6809 \\
\midrule
SYMPTOM & cz & \rank[1] & \rank[12] & \rank[6] & \rank[4] & \rank[12] & \rank[9] & 0.7360 & 0.5847 \\
SYMPTOM & en & \rank[1] & \rank[13] & \rank[8] & \rank[8] & \rank[14] & \rank[11] & 0.7791 & 0.6542 \\
SYMPTOM & es & \rank[3] & \rank[9] & \rank[6] & \rank[6] & \rank[9] & \rank[6] & 0.7914 & 0.6696 \\
SYMPTOM & it & \rank[1] & \rank[12] & \rank[5] & \rank[3] & \rank[12] & \rank[9] & 0.7637 & 0.5976 \\
SYMPTOM & nl & \rank[1] & \rank[15] & \rank[3] & \rank[4] & \rank[16] & \rank[9] & 0.7258 & 0.5637 \\
SYMPTOM & ro & \rank[1] & \rank[10] & \rank[3] & \rank[3] & \rank[9] & \rank[6] & 0.7743 & 0.6248 \\
SYMPTOM & sv & \rank[1] & \rank[10] & \rank[3] & \rank[3] & \rank[9] & \rank[5] & 0.7666 & 0.6187 \\\bottomrule
\end{tabular}
\caption{Full leaderboard summary for the Parallia submission across entity types and languages. Ranks are copied from the official leaderboard for each metric.}
\label{tab:appendix-full-results}
\end{table}

\section{Appendix: Training Dynamics}
\label{ann:training_loss}

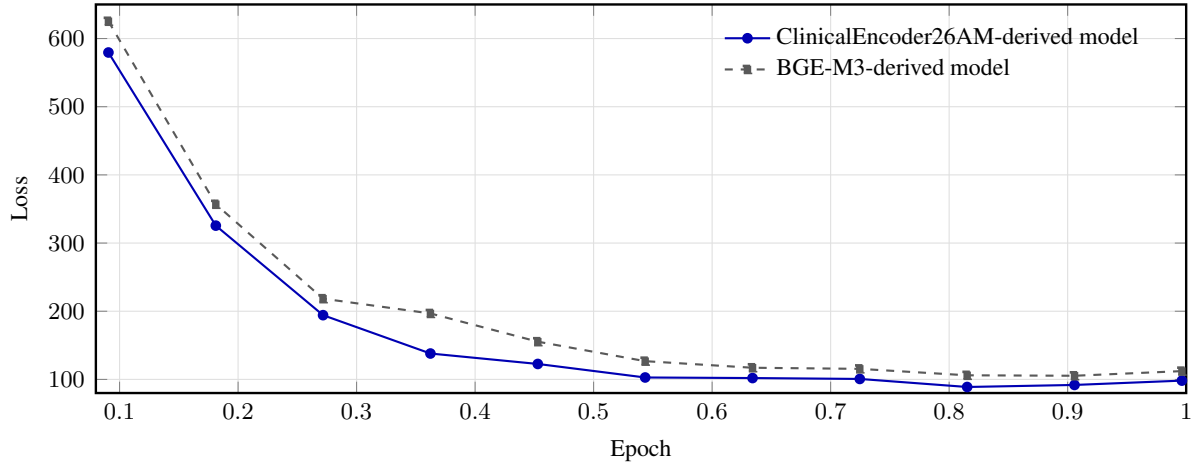
\begin{figure}[ht]
\centering
\begin{tikzpicture}
\begin{axis}[
    width=\textwidth,
    height=0.42\textwidth,
    xlabel={Epoch},
    ylabel={Loss},
    xmin=0.08,
    xmax=1.00,
    ymin=80,
    ymax=650,
    grid=both,
    major grid style={gray!25},
    minor grid style={gray!15},
    legend style={draw=none, fill=none, font=\small},
    legend cell align={left},
    legend pos=north east,
    tick label style={font=\small},
    label style={font=\small},
    line width=0.9pt,
]
\addplot[
    color=blue!70!black,
    thick,
    mark=*,
    mark size=1.6pt,
] coordinates {
    (0.0906,579.5365)
    (0.1812,325.5687)
    (0.2717,194.2215)
    (0.3623,137.9701)
    (0.4529,122.6525)
    (0.5435,102.8171)
    (0.6341,101.9777)
    (0.7246,100.6724)
    (0.8152,88.8760)
    (0.9058,91.8658)
    (0.9964,98.2115)
};
\addlegendentry{ClinicalEncoder26AM-derived model}

\addplot[
    color=black!65,
    thick,
    dashed,
    mark=square*,
    mark size=1.4pt,
] coordinates {
    (0.0906,625.3634)
    (0.1812,356.8888)
    (0.2717,218.4090)
    (0.3623,196.7429)
    (0.4529,155.3395)
    (0.5435,126.7580)
    (0.6341,117.0937)
    (0.7246,115.4627)
    (0.8152,106.0137)
    (0.9058,105.3420)
    (0.9964,112.1600)
};
\addlegendentry{BGE-M3-derived model}
\end{axis}
\end{tikzpicture}
\caption{Training loss over the first epoch for the ClinicalEncoder26AM-derived and BGE-M3-derived taggers. The clinical post-training initialization leads to faster convergence and lower loss through most of the run.}
\label{fig:training_loss}
\end{figure}

\end{document}